# A Joint Model of Language and Perception for Grounded Attribute Learning

Cynthia Matuszek · cynthia@cs.washington.edu
Nicholas FitzGerald · nfitz@cs.washington.edu
Luke Zettlemoyer · lsz@cs.washington.edu
Liefeng Bo · lfb@cs.washington.edu
Dieter Fox · fox@cs.washington.edu

Computer Science and Engineering, Box 352350, University of Washington, Seattle, WA 98195-2350

## Abstract

As robots become more ubiquitous and capable, it becomes ever more important for untrained users to easily interact with them. Recently, this has led to study of the *language grounding* problem, where the goal is to extract representations of the meanings of natural language tied to the physical world. We present an approach for joint learning of language and perception models for grounded attribute induction. The perception model includes classifiers for physical characteristics and a language model based on a probabilistic categorial grammar that enables the construction of compositional meaning representations. We evaluate on the task of interpreting sentences that describe sets of objects in a physical workspace, and demonstrate accurate task performance and effective latent-variable concept induction in physical grounded scenes.

## 1. Introduction

Physically grounded settings provide exciting opportunities for learning. For example, a person might be able to teach a robot about objects in its environment. However, to do this, a robot must jointly reason about the different modalities encountered (for example language and vision), and induce rich associations with as little guidance as possible.

Consider a simple sentence such as "These are the yellow blocks," uttered in a setting where there is a physical workspace that contains a number of objects that vary in shape and color. We assume that a robot can understand sentences like this if it can solve the associated *grounded object selection task*. Specifically, it must realize that words such as "yellow" and "blocks" refer to object attributes, and *ground* the meaning of such words by mapping them to a perceptual system that will enable it to identify the specific physical objects referred to. To do so robustly, even in cases where words or attributes are new, our robot must learn (1) visual classifiers that identify the appropriate object properties, (2) representations of the meaning of individual words that incorporate these classifiers, and (3) a model of compositional semantics used to analyze complete sentences.

In this paper, we present an approach for jointly learning these components. Our approach builds on existing work on visual attribute classification (Bo et al., 2011) and probabilistic categorial grammar induction for semantic parsing (Zettlemoyer & Collins, 2005; Kwiatkowski et al., 2011). Specifically, our system induces new grounded concepts (groups of words along with the parameters of the attribute classifier they are paired with) from a set of *scenes* containing only sentences, images, and indications of what objects are being referred to. As a result, it can be taught to recognize previously unknown object attributes by someone describing objects while pointing out the relevant objects in a set of training scenes. Learning is online, adding one scene at a time, and EM-like, in that the parameters are updated to maximize the expected marginal likelihood of the latent language and visual components of the model. This integrated approach allows for effective model updates with no explicit labeling of logical meaning representations or attribute classifier outputs.

We evaluate this approach on data gathered on Ama-zon Mechanical Turk, in which people describe sets of objects on a table. Experiments demonstrate that the joint learning approach can effectively extend the set of grounded concepts in an incomplete model initialized with supervised training on a small dataset. This provides a simple mechanism for learning vocabulary in a physical environment.

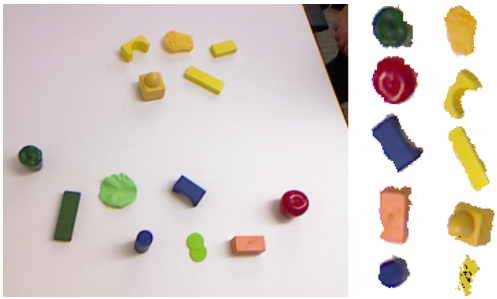

*Figure 1.* An example of an RGB-D object identification scene. Columns on the right show example segments, identified as positive (far right) and negative (center).

## 2. Overview of the Approach

**Problem** We wish to learn a joint language and perception model for the object selection task. The goal is to automatically map a natural language sentence $x$ and a set of scene objects $O$ to the subset $G \subseteq O$ of objects described by $x$. The left panel of Fig. 1 shows an example scene. Here, $O$ is the set of objects present in this scene. The individual objects $o \in O$ are extracted from the scene via segmentation (the right panel of Fig. 1 shows example segments). Given the sentence $x =$ "Here are the yellow ones," the goal is to select the five yellow objects for the named set $G$.

**Model Components** Given a sentence and segmented scene objects, we learn a distribution $P(G \mid x, O)$ over the selected set. Our approach combines recent models of language and vision, including:

(1) A *semantic parsing* model that defines $P(z|x)$, a distribution over logical meaning representations $z$ for each sentence $x$. In our running example, the desired representation $z = \lambda x.color(x, yellow)$ is a lambda-calculus expression that defines a set of objects that are yellow. For this task, we build on an existing semantic parsing model (Kwiatkowski et al., 2011).

(2) A set of *visual attribute classifiers* $C$, where each classifier $c \in C$ defines a distribution $P(c = true|o)$ of the classifier returning true for each possible object $o \in O$ in the scene. For example, there would be a unique classifier $c \in C$ for each possible color or shape an object can have. We use logistic regression to train classifiers on color and shape features extracted from object segments recorded using a Kinect depth camera.

**Joint Model** We combine these language and vision models in two ways. First, we introduce an explicit model of alignment between the logical constants in the logical form $z$ and classifiers in the set $C$. This alignment would, for example, enable us to learn that the logical constant *yellow* should be paired with a classifier $c \in C$ that fires on yellow objects.

Next, we introduce an execution model that allows us to determine what scene objects in $O$ would be selected by a logical expression $z$, given the classifiers in $C$. This allows us to, for example, execute $\lambda x.color(x, green) \wedge shape(x, triangle)$ by testing all of the objects with the appropriate classifiers (for *green* and *triangle*), then selecting objects on which both classifiers return true. This execution model includes uncertainty from the semantic parser $P(z|x)$, classifier confidences $P(c = true|o)$, and a deterministic ground-truth constraint that encodes what objects are actually intended to be selected. Full details are in Sec. 5.

**Model Learning** We present an approach that learns the meaning of new words from a dataset $D = \{(x_i, O_i, G_i) \mid i = 1 \ldots n\}$, where each example $i$ contains a sentence $x_i$, the objects $O_i$, and the selected set $G_i$. This setup is an abstraction of the situation where a teacher mentions $x_i$ while pointing to the objects $G_i \subseteq O_i$ she describes. As described in detail in Sec. 6, learning proceeds in an online, EM-like fashion by repeatedly estimating expectations over the latent logical forms $z_i$ and the outputs of the classifiers $c \in C$, then using these expectations to update the parameters of the component models for language $P(z|x)$ and visual classification $P(c|o)$. To bootstrap the learning approach, we first train a limited language and perception system in a fully supervised way: in this stage, each example additionally contains labeled logical meaning expressions and classifier outputs, as described in Sec. 6.

## 3. Related Work

To the best of our knowledge, this paper presents the first approach for jointly learning visual classifiers and semantic parsers, to produce rich, compositional models that span directly from sensors to meaning. However, there is significant related work on the model components, and on grounded learning in general.

**Vision** Current state-of-the-art object recognition systems (Felzenszwalb et al., 2009; Yang et al., 2009) are based on local image descriptors, for example SIFT over images (Lowe, 2004) and Spin Images over 3D point clouds (Johnson & Hebert, 1999). Visual

$$
\begin{array}{cccccccc}
\text{this} & \text{red} & \text{block} & \text{is} & \text{in the} & \text{shape} & \text{of a} & \text{half-pipe} \\
\hline
N/N & N & N\backslash N & S\backslash N/N & N/N & N/NP & NP/NP & NP \\
\lambda f.f & \lambda x.color(x, red) & \lambda f.f & \lambda f.\lambda g.\lambda x.f(x) \wedge g(x) & \lambda f.f & \lambda y.\lambda x.shape(x,y) & \lambda x.x & arch
\end{array}
$$

Figure 2. An example semantic analysis for a sentence from our dataset.

attributes provide rich descriptions of objects, and have become a popular topic in the vision community (Farhadi et al., 2009; Parikh & Grauman, 2011); although very successful, we still lack a deep understanding of the design rules underlying them and how they measure similarity. Recent work on kernel descriptors (Bo et al., 2010) shows that these hand-designed features are equivalent to a type of match kernel that performs similarly to sparse coding (Yang et al., 2009; Yu & Zhang, 2010) and deep networks (Lee et al., 2009) on many object recognition benchmarks (Bo et al., 2010). We adapt kernel descriptors as feature extractors for attribute classifiers because of their strong empirical performance.

**Semantic Parsing** There has been significant work on supervised learning for inducing semantic parsers (Zelle & Mooney, 1996; He & Young, 2006; Wong & Mooney, 2007). Our research builds on work on supervised learning of CCG parsers (Zettlemoyer & Collins, 2005; Kwiatkowski et al., 2011); there is also work on performing semantic analysis with alternate forms of supervision. Clarke (2010) and Liang (2011) describe approaches to learning semantic parsers from questions paired with database answers, while Goldwasser (2011) presents work on unsupervised learning. However, none of these approaches include joint models of language and vision.

**Grounding** There has been significant work on grounded learning more generally in the robotics and vision communities. A full review is beyond the scope of this paper, so we highlight a few examples. Roy developed a series of techniques for grounding words in visual scenes (Mavridis & Roy, 2006; Reckman et al., 2010; Gorniak & Roy, 2003). In computer vision, the grounding problem often relates to detecting objects and attributes in visual information (e.g., see (Barnard et al., 2003)); however, these approaches primarily focus on isolated word meaning, rather than compositional semantic analyses. Most closely related to our work are approaches that learn probabilistic language models from natural language input (Matuszek et al., 2012; Chen & Mooney, 2011), especially those that include a visual component (Tellex et al., 2011). However, these approaches ground language into predefined language formalisms, rather than extending the model to account for entirely novel input.

## 4. Background on Semantic Parsing

Our grounded language learning incorporates a state-of-the-art model, FUBL, for semantic parsing, as reviewed in this section. FUBL (Kwiatkowski et al., 2011) is an algorithm for learning factored Combinatory Categorial Grammar (CCG) lexicons for semantic parsing. Given a dataset $\{(x_i, z_i) \mid i = 1...n\}$ of natural language sentences $x_i$, which are paired with logical forms $z_i$ that represent their meaning, UBL learns a factored lexicon $\Lambda$ made up of a set of lexemes $L$ and a set of lexical templates $T$. Lexemes combine with templates in order to form lexical items, which can be used by a semantic parser to parse natural language sentences into logical forms. For example, given the sentence $x = $"this red block is in the shape of a half-pipe" and the logical form $z_i = \lambda x.color(x, red) \wedge shape(x, arch)$, FUBL learns a parse like the example in figure 2. In this parse, the lexeme (half-pipe, [arch]) has combined with the template $\lambda(\omega, \vec{v}).[\omega \vdash NP : \vec{v}_1]$ to yield the lexical item $half\text{-}pipe \vdash NP : arch$.

FUBL also learns a log-linear model which produces the probability of a parse $y$ that yields logical form $z$ given the sentence $x$:

$$P(y, z \mid x; \Theta^L, \Lambda) = \frac{e^{\Theta^L \cdot \phi(x,y,z)}}{\sum_{(y', z')} e^{\Theta^L \cdot \phi(x,y',z')}} \quad (1)$$

where $\phi(x, y, z)$ is a feature vector encompassing the lexemes and lexical templates used to generate $y$, amongst other things.

In this work, we initialize our parse model using the standard FUBL approach, followed by automatically inducing lexemes paired with new visual attributes not present in the initial training set, as we will see in the next section.

## 5. Joint Language/Perception Model

As described in Sec. 2, the object selection task is to identify a subset of objects, $G$, given a scene $O$ and an NL sentence $x$. We define a *possible world* $w$ to be a set of classifier outputs, where $w_{o,c} \in \{T, F\}$ specifies the boolean output of classifier $c$ for object $o$. Our joint probabilistic model is:

$$P(G \mid x, O) = \sum_z \sum_w P(G, z, w \mid x, O) \qquad (2)$$

where the latent variable $z$ over logical forms models linguistic uncertainty and the latent $w$ over possible worlds models perceptual uncertainty.

We further decompose (2) into a product of models for language, vision, and grounded execution. This final model selects the named objects $G$, motivated in Sec. 2 and described below; the final decomposition is:

$$P(G, z, w \mid x, O) = P(z \mid x) P(w \mid O) P(G \mid z, w) \qquad (3)$$

Here, the language model $P(z|x)$ and vision model $P(w|O)$ are held in agreement by the conditional probability term $P(G|z,w)$. Let $z(w)$ be the set of objects that are selected, under the assignment in $w$, when $z$ is applied to them. For example, the expression $z = \lambda x.shape(x, cube) \wedge color(x, red)$ would return true when applied to the objects in $w$ for which the classifiers for the *cube* and *red* logical constants return true. Now, $P(G|z,w)$ forces agreement and models object selection by putting all of its probability mass on the set $G$ that equals $z(w)$.

In this formulation, the language and vision distributions are conditionally independent given this agreement. The semantic parsing model $P(z|x)$ builds on previous work, as described in eqn. (1). The perceptual classification $P(w|O)$ is defined as follows: we assume each perceptual classifier is applied independently, decomposing this term into:

$$P(w \mid O) = \prod_{o \in O} \prod_{c \in C} P(w_{o,c} | o) \qquad (4)$$

where the probability of a world is simply the product of the probabilities of the individual classifier assignments for all of the objects.

Each classifier is a logistic regression model, where the probability of a classifier $c$ on a given object $o$ is:

$$P(w_{o,c} = 1 | o; \Theta^P) = \frac{e^{\Theta_c^P \cdot \phi(o)}}{1 + e^{\Theta_c^P \cdot \phi(o)}} \qquad (5)$$

where $\Theta_c^P$ is the parameters in $\Theta^P$ for classifier $c$. This approach provides a simple, direct way to couple the individual language and vision components to model the object selection task.

**Inference** There are two key inference problems in a model of this type. During learning, we need to compute the marginal distribution $P(z, w|x, O, G)$ over latent logical forms $z$ and perceptual assignments $w$ (see next section). At test time, we must compute $\arg\max_G P(G|x, O)$ to find the set of named objects.

Computing this probability distribution requires summing the total probability of all world/logical form pairs that name $G$. For each possible world $w$, determining if $z$ names $G$ is equivalent to a SAT problem, as $z$ can theoretically encode an arbitrary logical expression that will name the appropriate $G$ only when satisfied. Computing the marginal probability is then a weighted model counting problem, which is in #-P. However, the logical expressions allowed by our current grammar—conjunctions of unary attribute descriptors—admit efficient exact computation, described below.

## 6. Model Learning

The physically grounded joint learning problem is to induce a model $P(G|x, O)$, given data of the form $D = \{(x_i, O_i, G_i) \mid i = 1 \ldots n\}$, where each example $i$ contains a sentence $x_i$, the objects $O_i$, and the selected set $G_i$. We consider the case where the learner already has a partial model, including a CCG parser with a small vocabulary and a small set of attribute classifiers. The goal is to automatically extend the model to induce new classifiers that are tied to new words in the semantic parser. We first describe the learning algorithm, then present how we initialize the approach by learning decoupled models from small datasets with more extensive annotations.

**Aligning Words to Classifiers** One key challenge is to learn to create new attribute classifiers associated with unseen words in the sentences $x_i$ in the data $D$. We take a simple, exhaustive approach by creating a set of $k$ new classifiers, initialized to uniform distributions. Each classifier is additionally paired with a new logical constant in the FUBL lambda-calculus language. Finally, a new lexeme is created by pairing each previously unknown word in a sentence in $D$ with either one of these new classifier constants, or the logical expressions from an existing lexeme in the lexicon. The parsing weights for the indicator features for each of these additions are set to 0. This approach learns, through the probabilistic updates described below, to jointly reestimate the parameters of both the new classifiers and the expanded semantic parsing model.

**Parameter Estimation** We aim to estimate the language parameters $\Theta^L$ and perception parameters

$\Theta^P$ from data $D = \{(x_i, O_i, G_i) \mid i = 1 \ldots n\}$, as defined above. We want to find parameter settings that maximize the marginal log likelihood of $D$:

$$LL(D; \Theta^L, \Theta^P) = \sum_{i=1\ldots n} \ln P(G_i | x_i, O_i; \Theta^L, \Theta^P) \quad (6)$$

This objective is non-convex due to the sum over latent assignments for the logical form $z$ and attribute classifier outputs $w$ in the definition of $P(G_i|x_i, O_i; \Theta^L, \Theta^P)$ from eqn. (2). However, if $z$ and $w$ are labeled, the overall algorithm reduces to simply training the log-linear models for the semantic parser $P(z|x_i; \Theta^L)$ and attribute classifiers $P(w|O_i; \Theta^P)$, both well-studied problems. In this situation, we can use an EM algorithm to first estimate the marginal $P(z, w \mid x_i, O_i, G_i; \Theta^L, \Theta^P)$, then maximize the expected likelihood according to the distribution, with a weighted version of our familiar log-linear model parameter updates. We present an online version of this approach, with updates computed one example at a time.

*Computing Expectations* For each example $i$, we must compute the marginal over latent variables given by:

$$P(z, w \mid x_i, O_i, G_i; \Theta^L, \Theta^P) = \frac{P(z \mid x_i; \Theta^L) P(w \mid O_i; \Theta^P) P(G_i|z, w)}{\sum_{z'} \sum_{w'} P(z' \mid x_i; \Theta^L) P(w' \mid O_i; \Theta^P) P(G_i|z', w')} \quad (7)$$

Since computing all possible parses $z$ is exponential in the length of the sentence, we use beam search to find the top-N parses. This exact inference could be replaced with an approximate method, such as MC-SAT, to accommodate a more permissive grammar.

*Conditional Expected Gradient* For each example, we update the parameters with the expected gradient, according to the marginal distribution above. For the language parameters $\Theta^L$, the gradient is

$$\Delta^L = \sum_{z'} \sum_{w'} P(z', w' \mid x_i, O_i, G_i; \Theta^L, \Theta^P) * \\ (E_{P(y|x_i, z'; \Theta^L)} \left[\phi_j^L(x_i, y, z')\right] - \\ E_{P(y,z|x_i; \Theta^L)} \left[\phi_j^L(x_i, y, z)\right]) \quad (8)$$

where the inner difference of expectations is the familiar gradient of a log-linear model for conditional random fields with hidden variables (Quattoni et al., 2007; Kwiatkowski et al., 2010), and is weighted according to the expectation.

Similarly, for the perception parameters $\Theta^P$, the gradient is:

$$\Delta_c^P = \sum_{z'} \sum_{w'} P(z', w' \mid x_i, O_i, G_i; \Theta^L, \Theta^P) * \\ \sum_{o \in O_i} \left[w'_{o,c} - P(w'_{o,c} = 1 \mid \phi(o); \Theta^P)\right] \phi(o) \quad (9)$$

where the inner sum ranges over the objects and adds in the familiar gradient for logistic regression binary-classification models.

*Online Updates* We use a simple, online parameter estimation scheme that loops over the data $K = 10$ (picked on validation set) times. For each data point $i$ consisting of the tuple $(x_i, O_i, G_i)$, we perform an update where we take a step according to the above expected gradient over the latent variables. We use a learning rate of 0.1 with a constant decay of .00001 per update for all experiments.

**Discussion** This complete learning approach provides an efficient online algorithm that closely matches the style of interactive, grounded language learning we are pursuing in this work. Given the decayed learning rate, the algorithm is guaranteed to converge, but little can be said about the optimality of the solution. However, as we see in Sec. 7, the approach works well in practice for the object set selection task we consider.

**Bootstrapping** To construct the initial limited language and perceptual models, we make use of a small, supervised data set $D_{sup} = \{(x_i, z_i, w_i, O_i, G_i) \mid i = 1 \ldots m\}$, which matches our previous setup but additionally labels the latent logical form $z_i$ and classifier outputs $w_i$. As mentioned above, learning in this setting is completely decoupled and we can estimate the semantic parsing distribution $P(z_i|x_i; \Theta^L)$ with the FUBL learning algorithm (Kwiatkowski et al., 2011) and the attribute classifiers $P(w_i|O_i; \Theta^P)$ with gradient ascent for logistic regression. As we show experimentally, $D_{sup}$ can often be quite small, and will in general not contain many of the words and attributes that must be additionally learned in the full approach. Exploring approaches for learning without $D_{sup}$, such as replacing it with interactive dialog with a human teacher, is an important area for future work.

## 7. Experimental Setup

**Data Set** Data was collected using a selection of toys, including wooden blocks, plastic food, and building bricks. For each scene, we collected short RGB-D videos with a Kinect depth camera, showing a person gesturing to a subset of the objects. Natural language annotations were gathered using Mechanical Turk; workers were asked to describe the objects being pointed to in the video (see Fig. 3). The referenced objects were then marked as belonging to $G$, the positive set of objects for that scene. A total of 142 scenes were shown, eliciting descriptions of 12 attributes, divided evenly into shapes and colors. In total, there were 1003 sentence/annotation pairs.

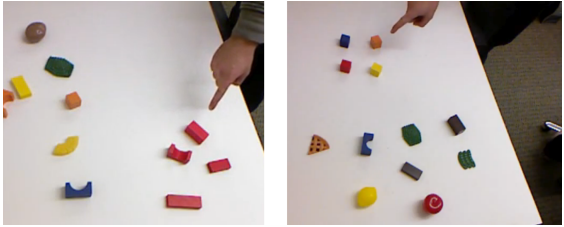

*Figure 3.* Example scenes presented on Mechanical Turk. Left: A scene that elicited the descriptions "here are some red things" and "these are various types of red colored objects", both labeled as $\lambda x.color(x, red)$. Right: A scene associated with sentence/meaning pairs such as "this toy is orange cube" and $\lambda x.color(x, orange) \wedge shape(x, cube)$.

**Perceptual Features** To automatically segment objects from each scene, we performed RANSAC plane fitting on the Kinect depth values to find the table plane, then extracted connected components (segments) of points more than a minimum distance above that plane. After getting segmented objects, features for every object are extracted using kernel descriptors (Bo et al., 2011). We extract two types of features, for depth values and RGB values; these correspond to shape and color attributes, respectively. During training, the system learns logistic regression classifiers using these features. In the initialization phase used to bootstrap the model, the annotation provides information about which language attributes relate to shape or color. However, this information is not provided in the training phase.

**Language Features** We follow (Kwiatkowski et al., 2011) in including a standard set of binary indicator features to define the log-linear model $P(z|x; \Theta^L)$ over logical forms, given sentences. This includes indicators for which lexical entries were used and properties of the logical forms that are constructed. These features allow the joint learning approach to weight lexical selection against evidence provided by the compositional analysis and the visual model components.

## 8. Results

This section presents results and a discussion of our evaluation. We demonstrate effective learning in the full model for the object set selection task. We then briefly describe ablation studies and examples of learned models..

### 8.1. Object Set Selection

To measure set selection task performance, we divided the data according to attribute. To initialize the model, we used the data for six of the attributes to train supervised classifiers, and provided logical forms for the corresponding sentences to train the initial semantic parsing model, as described at the end of Sec. 6. Data for the remaining six attributes were used for evaluation, with 80% allocated for training and 20% held out for testing. Here, all of the visual scenes are previously unseen, the words in the sentences describing the new attributes are unknown, and the only available labels are the output object set $G$.

We report precision, recall, and F1-score on the set selection task. Results are averaged over 10 different runs with the training data presented in different randomized orders. The system performs well, achieving an average precision of 82%, recall of 71%, and a 76% F1-score. This level of performance is achieved relatively quickly; performance generally converges within five passes over the training data.

### 8.2. Ablation Studies

To examine the need for a joint model, we measure performance of two models in which either the language or the visual component is sharply limited. In each case, performance significantly degrades. These results are summarized in Fig. 4.

**Vision** In order to measure how a set of classifiers would perform on the set selection task with only a simple language model, we manually created a thesaurus of words used in the dataset to refer to target attributes containing, on average, 5 different ways of referring to each color and shape. To learn the unsupervised concepts for this baseline, we first extracted a list of all words appearing in the training corpus but not in the initialization data; words which appear in the thesaurus are grouped into *synonym sets*. To train classifiers, we collect objects from scenes in which only terms from the given synonym set appear. Any synonym set which does not occur in at least 2 distinct scenes is discarded. The resulting positive and negative objects are used to train classifiers. To generate a predicted set of objects at test time, we find all synonym sets which occur in the sentence $x$, and determine whether the classifiers associated with those words successfully identify the object.

Averaged across our trials, the results are as follows: Precision=0.92; Recall=0.41; F1-score=0.55. These results are, on average, notably worse than the performance of the jointly trained model.

**Semantic Parsing** As a baseline for testing how well a pure parsing model will perform when the perception model is ablated, we run the parsing model obtained during initialization directly on the test set, training no new classifiers. Since the parser is capable of generating parses by skipping unknown words, this baseline is equivalent to treating the unknown concept words as if they are semantically empty.

Averaged across our trials, the results are as follows: Precision=0.52; Recall=0.09; F1-score=0.14. Not surprisingly, a substantial number of parses selected no objects, as the parser has no way of determining the meaning of an unknown word.

|  | Precision | Recall | F1-Score |
|---|---|---|---|
| Vision | **0.92** | 0.41 | 0.55 |
| Language | 0.52 | 0.09 | 0.14 |
| **Joint** | 0.82 | **0.71** | **0.76** |

*Figure 4.* A summary of precision, recall, and F1 for ablated models and the joint learning model.

### 8.3. Discussion and Examples

This section discusses typical training runs and data requirements. We present examples of learned models, highlighting what is learned and typical errors, and then describe simple experiment investigating the amount of supervised data required for initialization.

Classifier performance after training effects the system's ability to perform the set selection task. During a sample trial, average accuracy of color and shape classifiers for newly learned concepts are 97% and 74%, respectively. Although these values are sufficient for reasonable task performance, there are some failures—for example, the shape attributes "cube" and "cylinder" are sometimes challenging to differentiate.

As noted in Sec. 4, the semantic parser contains lexemes that pair words with learned classifiers, and features that indicate lexeme use during parsing. Fig. 5 shows some selected word/classifier pairs, along with the weight for their associated feature (each trial produces a large number of such lexemes). The classifiers NEW0–NEW2 and NEW3–NEW5 are color and shape classifiers, respectively. As can be seen, each of the novel attributes is most strongly associated with a newly-created classifier, while irrelevant words such as "thing" tend to parse to null. The system must identify which of the classifier types to use for novel words.

We ran additional tests investigating whether the system is able to learn synonyms. Here, we split the data so that the training set has attributes learned during initialization, but are referred to by new, synonymous words. These runs performed comparably to those reported above; the approach easily learns lexemes that pair these new words with the appropriate classifiers.

Finally, we briefly discuss the effects of reducing the amount of annotated data used to initialize the language and perception model (see Fig. 6). As can be seen, with fewer than 150 sentences, the learned grammar does not seem to have sufficient coverage to model unknown words in joint learning; however, beyond that, performance is quite stable.

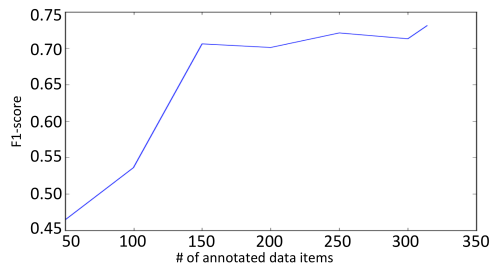

|  | NEW0 | NEW1 | NEW2 | NEW3 | NEW4 | NEW5 | null |
|---|---|---|---|---|---|---|---|
| "red" | 3.27 | -0.34 | -0.37 | -0.16 | -0.16 | -0.17 | 0.00 |
| "green" | -0.39 | -0.30 | 3.47 | -0.19 | -0.19 | -0.19 | 0.00 |
| "blue" | -0.34 | 2.97 | -0.31 | -0.16 | -0.16 | -0.16 | 0.00 |
| "thing" | 0.00 | 0.00 | 0.00 | 0.00 | 0.00 | 0.00 | 0.29 |
| "cube" | -0.43 | 0.31 | -0.37 | -0.23 | 0.00 | 2.78 | 0.00 |
| "that" | 0.00 | 0.00 | 0.00 | 0.00 | 0.00 | 0.00 | 0.42 |
| "arch" | -0.01 | -0.01 | 0.09 | -0.14 | 0.6 | -0.15 | 0.00 |
| "triangle" | 0.34 | -0.30 | 0.04 | 1.92 | -0.18 | -0.19 | 0.00 |
| "toys" | 0.00 | 0.00 | 0.00 | 0.00 | 0.00 | 0.00 | 0.38 |

*Figure 5.* Feature weights for hypothesized lexemes pairing natural language words (rows) with newly created terms referring to novel classifiers (columns), as well as the special null token. Each weight serves as an unnormalized indicator of which associations are preferred.

*Figure 6.* Example F1-score on object recognition from models initialized with reduced amounts of labeled data, reported over one particular data split. The F1-score for this split peaks at roughly 73%.

## 9. Conclusion

This paper presents a joint model of language and perception for grounded attribute learning. Our approach learns representations of the meanings of natural language, using visual perception to ground those meanings in the physical world. Learning is performed via optimizing the data log-likelihood using an online, EM-like training algorithm. This system is able to learn accurate language and attribute models for the object set selection task, given data containing only language, raw percepts, and the target objects. By jointly learning language and perception models, the approach can identify which novel words are color attributes, shape attributes, or no attributes at all.

We believe our approach has significant potential to scale to general language grounding problems. In particular, our modular framework was designed to easily incorporate future advances in visual classification and semantic parsing. We are also working to scale the complexity of the language and physical scenes, with the eventual goal of robust learning in completely unconstrained environments.

# Acknowledgments

This work was funded in part by the Intel Science and Technology Center for Pervasive Computing, the Robotics Consortium sponsored by the U.S. Army Research Laboratory under the Collaborative Technology Alliance Program (W911NF-10-2-0016), and NSF grant IIS-1115966.